\documentclass{article} 
\usepackage{main,times}

\usepackage{amsmath,amsfonts,bm}

\def\eqref#1{equation~\ref{#1}}

\def\1{\bm{1}}

\DeclareMathAlphabet{\mathsfit}{\encodingdefault}{\sfdefault}{m}{sl}
\SetMathAlphabet{\mathsfit}{bold}{\encodingdefault}{\sfdefault}{bx}{n}

\definecolor{Coral}{rgb}{1, 0.47, 0.24}

\newcommand{\boldparagraph}[1]{\par\vspace{0.2em}\noindent{\bf #1.}}

\usepackage{booktabs}
\usepackage{multirow}
\usepackage{graphicx}
\usepackage{array}
\usepackage{gensymb}  
\usepackage{fancyvrb}
\usepackage{lipsum}
\usepackage{xspace}
\usepackage{pifont}

\usepackage{hyperref}
\usepackage{url}
\usepackage{graphicx}
\usepackage{rotating}
\usepackage{subfig}
\usepackage{booktabs}
\usepackage{threeparttable}
\usepackage{makecell}
\usepackage{changepage}

\title{GeoVerse: World-Consistent Novel View \\  Synthesis in Geometric Latent Space}

\author{Kerui Ren$^{1,2}$ \quad 
Tao Lu$^{2}$ \quad
Linning Xu$^{3}$ \quad
Changjian Jiang$^{4}$ \quad \\
\textbf{Mu Huang}$^{5}$ \quad
\textbf{Chunhua Shen}$^{6,2}$ \quad
\textbf{Mulin Yu}$^{2}$\footnotemark[2] \quad
\textbf{Bo Dai}$^{4}$\footnotemark[2] \quad \\
{\small$^1$Shanghai Jiao Tong University, $^2$Shanghai Artificial Intelligence Laboratory, } \\
{\small$^3$The Chinese University of Hong Kong, $^4$The University of Hong Kong, } \\
{\small$^5$Fudan University,  $^6$Zhejiang University} \\
}

\iclrfinalcopy
\begin{document}

\maketitle
\let\thefootnote\relax\footnotetext{$^{\dagger}$Corresponding authors.}

\begin{figure*}[htbp]
\centering
\includegraphics[width=\linewidth]{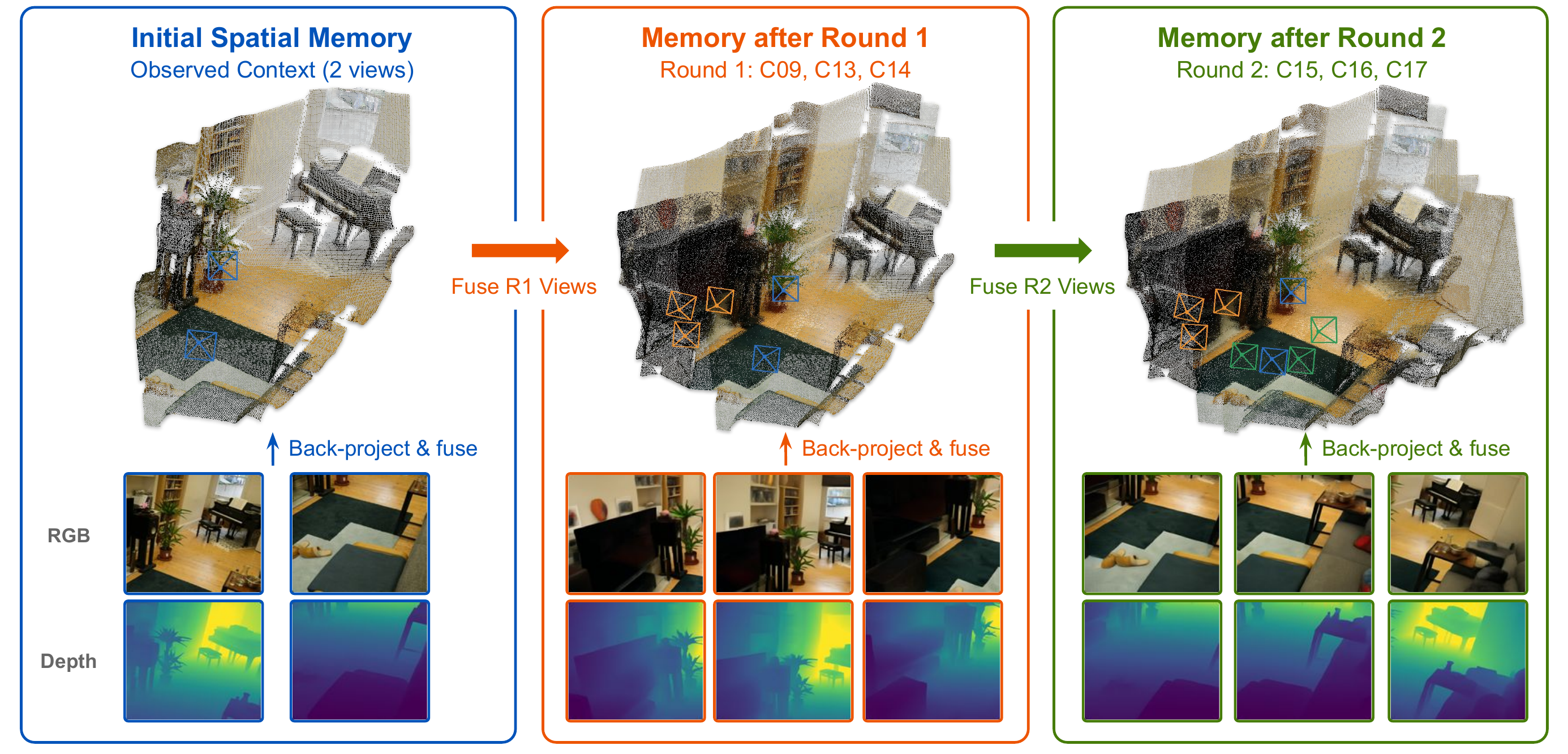}
\caption{GeoVerse enables long-sequence novel view synthesis by iteratively integrating generated observations into a persistent spatial memory, which in turn guides subsequent view synthesis. Project page: 
\href{https://geoverse-nvs.github.io/}{\textcolor{magenta}{\textbf{https://geoverse-nvs.github.io/}}}.}
\label{fig:teaser}
\end{figure*}

\begin{abstract}
Novel view synthesis from sparse images must reconcile faithful reconstruction of observed regions with plausible completion of unseen content, while maintaining world consistency across viewpoints.
Existing geometry-based methods preserve observed scene structure but often struggle to complete unseen regions, whereas video generative models offer rich appearance priors but accumulate inconsistencies during sequential view generation.
We propose GeoVerse, a framework that synthesizes world-consistent novel views by performing generation within the geometric latent space of a pretrained 3D foundation model and injecting appearance priors from a video generative model.
Specifically, GeoVerse extracts multilevel features from Wan2.2 VACE and injects them into the geometric latent diffusion model via a ControlNet-style adapter, incorporating video-learned appearance priors to enhance structural completion.
To enforce cross-view coherence, a global spatial memory continuously aggregates observed and synthesized content, reprojecting target-aligned guidance to anchor subsequent predictions to a shared scene representation.
Extensive experiments across diverse datasets demonstrate improved visual quality and geometric consistency, with a 2.23~dB higher PSNR on DL3DV and 32.4\% lower ATE on Mip-NeRF360 compared to GLD.
\end{abstract}

\section{Introduction}
\label{sec:intro}

Novel View Synthesis (NVS) is a cornerstone task in 3D computer vision that aims to render photorealistic images from specified, previously unseen target viewpoints given one or more reference images~\citep{nerf2020,viewcrafter2024}.
A fundamental challenge in NVS lies in balancing reconstruction fidelity during view interpolation with generative capability during view extrapolation, faithfully aggregating observed scene content while plausibly completing unobserved regions.
To bridge faithful reconstruction with generative completion, maintaining a unified geometric representation is crucial for ensuring spatial and semantic coherence across shifting viewpoints.

Early NVS frameworks were predominantly constrained to per-scene optimization, with pioneering paradigms like Neural Radiance Fields (NeRF)~\citep{nerf2020} and 3D Gaussian Splatting (3DGS)~\citep{kerbl2023} fitting scene-specific representations on dense captures.
To bypass scene-specific training bottlenecks, recent generalizable architectures leverage scaled data to predict Gaussian primitives directly from sparse views in a feed-forward manner~\citep{mvsplat2024,anysplat2025}.
Spurred by 3D foundation models~\citep{dust3r2023,vggt2025}, these feed-forward methods achieve rapid feed-forward reconstruction with strong geometry grounding~\citep{anysplat2025,worldmirror2025}.
However, bound by deterministic geometric formulations, pure reconstruction frameworks inherently lack generative imagination, frequently producing severe visual artifacts, blurriness, or hollow voids when synthesizing unobserved regions under large camera motions.

To compensate for the limited extrapolation capability of reconstruction methods, video diffusion models have recently been repurposed for view synthesis~\citep{viewcrafter2024}, leveraging rich appearance priors learned from vast video corpora~\citep{wan2025}.
Despite their expressive completion, applying video models directly or sequentially to 3D scenes often accumulates cross-frame inconsistencies, causing drift and structural breakdown over long horizons~\citep{spatialmemory2025,latentspatialmemory2026}.
To enforce spatial consistency, hybrid frameworks align video diffusion features with 3D representations~\citep{geometryforcing2025,gen3r2026}, yet they remain burdened by the heavy computational overhead of synthesizing dense video frames. More recently, Geometry Latent Diffusion (GLD)~\citep{gld2026} has emerged as a promising alternative that generates novel views directly within a geometric latent space. Built upon 3D foundation models such as DA3~\citep{da3_2025}, which is pretrained on large-scale data with depth and camera-pose supervision to learn cross-view attention, this space provides a robust structural foundation and a geometric prior for single-pass consistency. Furthermore, GLD's RGB head effectively decodes fine texture and appearance details from these features~\citep{gld2026}. However, restricted by a limited training distribution, GLD struggles with generative fidelity and stability on in-the-wild scenes.

In summary, existing NVS approaches struggle to seamlessly harmonize rigid 3D geometric consistency with expressively detailed generative completion.
To bridge this gap, we present \textbf{GeoVerse}, a novel framework that endows geometric latent diffusion with rich video generative priors while maintaining structural consistency via a persistent spatial memory.
We address these limitations through three key designs. First, we incorporate Wan2.2 VACE~\citep{wan2025,vace2025}, a high-capacity video diffusion model pretrained on extensive video data, to enrich the geometric latent space with powerful appearance priors and plausible scene completion. Second, we scale the 3D training corpus from 4 to 15 diverse real and synthetic datasets to broaden scene coverage and enhance cross-domain generalization.
Beyond single-pass consistency, successive view expansion requires a persistent representation across multiple inference steps. Inspired by ViewCrafter~\citep{viewcrafter2024}, we maintain a global colored point-cloud memory that accumulates historical content. Reprojecting this memory into target views provides pixel-aligned guidance, anchoring multi-step predictions while facilitating unobserved region completion. Leveraging this structured guidance, we employ reflow distillation~\citep{yan2024perflow} to reduce the denoising process to 4 steps, cutting per-inference latency from over 100 seconds to under 10 seconds. 

Our primary contributions are summarized as follows:\begin{itemize}
\item We introduce a novel framework that seamlessly bridges geometric latent diffusion with expressive video generative priors for world-consistent novel view synthesis.
\item We propose a global spatial memory mechanism that provides target-aligned input guidance, effectively enforcing sustained long-sequence geometric consistency across extended trajectories without cumulative drift.
\item We scale up model training across diverse real and synthetic multi-view corpora to significantly enhance cross-domain generalization and integrate a few-step reflow distillation scheme to significantly accelerate inference to under 10 seconds.
\end{itemize}

\section{Related Work}
\label{sec:related}

\subsection{Reconstructive Novel View Synthesis}

Focusing on fusing existing scene content within captured view bounds, reconstructive novel view synthesis was initially dominated by scene-specific optimization. Pioneered by NeRF \citep{nerf2020}, implicit radiance fields achieved novel view rendering through differentiable volume rendering, which was later accelerated and scaled by subsequent variants \citep{mipnerf2021,instantngp2022}. To overcome the implicit rendering bottleneck, 3DGS~\citep{kerbl2023} introduced explicit 3D Gaussian primitives for real-time synthesis, followed by improvements in anti-aliasing and geometry modeling \citep{mipsplatting2024,scaffoldgs2024,octreegs2024}. While capable of rendering high-fidelity views, these per-scene representations suffer from long optimization times and strictly require dense multi-view coverage. To bypass per-scene optimization, feed-forward frameworks learn generalizable mappings from sparse inputs to novel target views. Early generalizable architectures like MVSplat \citep{mvsplat2024} leverage plane-sweep cost volumes for 3D Gaussian prediction, while LVSM \citep{lvsm2024} scales transformer-based synthesis with minimal explicit geometric priors. Accelerated by 3D foundation models like VGGT \citep{vggt2025}, recent frameworks seamlessly unify camera estimation and feed-forward reconstruction. Specifically, AnySplat \citep{anysplat2025} jointly recovers camera poses and 3D Gaussians from unconstrained views, whereas WorldMirror \citep{worldmirror2025} incorporates multi-source geometric priors. In the monocular regime, SHARP \citep{sharp2026} regresses a metric 3D Gaussian field from a single image in one forward pass. Although feed-forward methods excel at interpolative synthesis within bounded views, extrapolating to unobserved regions remains inherently ambiguous.

\subsection{Generative Novel View Synthesis}

Generative novel view synthesis introduces generative priors to synthesize content beyond the observed scene coverage. ViewCrafter combines point-based 3D clues with a pretrained video diffusion model and iteratively expands both its camera trajectory and reconstructed content \citep{viewcrafter2024}. NeoVerse couples feed-forward 4D reconstruction with novel-trajectory video generation to model scenes from in-the-wild monocular videos \citep{neoverse2026}. To reduce drift over longer horizons, spatial-memory approaches explicitly store and retrieve previously generated scene content: geometry-grounded long-term memory caches 3D history, while Mirage lifts diffusion latents into a persistent 3D cache and queries it through latent-space warping \citep{spatialmemory2025,latentspatialmemory2026}. Lyra 2.0 combines per frame geometric matching for history retrieval with self augmented history training to suppress cumulative drift during long sequence generation~\citep{shen2026lyra2}. These methods improve long-horizon consistency while retaining a video-based generation pipeline. 

Geometric foundation models offer a complementary route to geometry-aware extrapolation. Geometry Forcing supervises intermediate video-diffusion representations with features from a geometric foundation model, while Gen3R adapts VGGT tokens into geometric latents and aligns them with pretrained video appearance latents for joint RGB and 3D generation \citep{geometryforcing2025,gen3r2026}. Although these methods improve the 3D awareness of video diffusion, their generation process remains organized as a video sequence. Geometric Latent Diffusion (GLD) instead repurposes the geometric feature space of DA3 as the native latent space for multi-view diffusion, enabling direct generation at specified target cameras with strong cross-view correspondence \citep{da3_2025,gld2026}. Unlike prior sequence-based approaches, GeoVerse adopts the geometric-latent formulation, augmented by video generative priors and a persistent 3D memory, enabling direct target-view synthesis free of dense video interpolation.
\section{Method}
\label{sec:method}

\begin{figure*}[t]
    \centering
    \includegraphics[width=\textwidth]{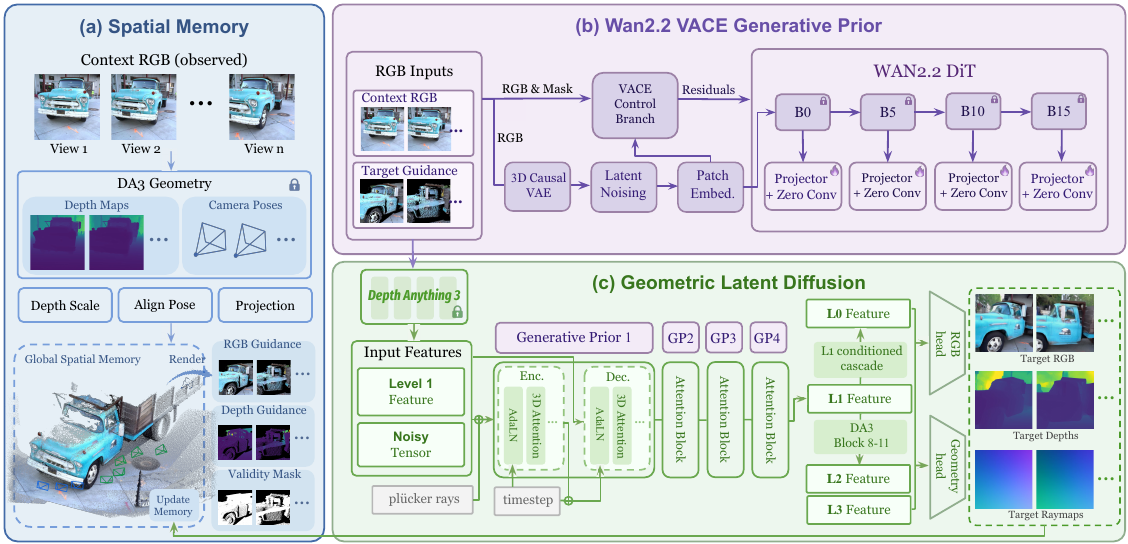}
    \caption{\textbf{Overview of GeoVerse.}
    Context observations initialize a global spatial memory that provides target-aligned RGB-D hints.
    Guided by the hints and Wan2.2 VACE features injected through a ControlNet-style adapter, geometric latent diffusion synthesizes target views in four denoising updates.
    The decoded RGB and geometry update the memory for subsequent view expansion.}
    \label{fig:geoverse_method}
\vspace{-4pt}
\end{figure*}

Fig.~\ref{fig:geoverse_method} illustrates the overall pipeline of GeoVerse, a framework that integrates geometric latent diffusion with video generative priors and a global spatial memory for world-consistent novel view synthesis. Let $\mathcal C$ and $\mathcal T$ denote the context and target view sets, with sizes $N_{\mathcal C}$ and $N_{\mathcal T}$, respectively. Given context images $\mathbf I_{\mathcal C}$ and target-aligned guidance $\mathbf I_{\mathcal T}^{\mathrm{proj}}$, GeoVerse predicts target RGB images $\hat{\mathbf I_{\mathcal T}}$ and geometry $\hat{\mathbf G_{\mathcal T}}=\{\hat{\mathbf D_{\mathcal T}},\hat{\mathbf R_{\mathcal T}}\}$, comprising depth and raymaps, and then updates the spatial memory $\mathbf M$. Specifically, Sec.~\ref{sec:method_prior} details the integration of geometric latent diffusion with video generative priors, Sec.~\ref{sec:method_memory} describes the spatial-memory aggregation and target-aligned projection, and Sec.~\ref{sec:method_efficiency} outlines the model distillation strategy for efficient inference.

\subsection{Enhancing Geometric Latent Diffusion with Video Priors}
\label{sec:method_prior}
\boldparagraph{Geometry Latent Space}
Standard image-video latent spaces provide robust appearance priors but struggle to maintain explicit 3D correspondence across novel viewpoints~\citep{rombach2022}. In contrast, GLD combines the geometry head of Depth Anything 3~\citep{da3_2025} for spatial layouts (depth and raymaps) with an auxiliary RGB head for high-frequency textures~\citep{gld2026}. Following GLD~\citep{gld2026}, we perform diffusion generation directly within the multi-level geometric latent space $\mathcal{F}=\{\mathbf{F}^i\}_{i=0}^{3}$ of a frozen DA3 encoder, where feature levels correspond to Transformer blocks $(b_0,b_1,b_2,b_3)=(5,7,9,11)$, and designate Level-1 as the synthesis boundary to balance geometric accuracy and visual fidelity. Unlike GLD's zero-padding target conditions, GeoVerse constructs the input sequence $\overline{\mathbf I}=[\mathbf I_{\mathcal C},\mathbf I_{\mathcal T}^{\mathrm{proj}}]$ by merging context images with spatial memory projections, which are then processed by the frozen DA3 encoder up to the synthesis boundary:
\begin{equation}
\overline{\mathbf F}
=\mathbf V^{b_1}\odot\mathcal E_{\mathrm{geo}}^{1:b_{1}}(\overline{\mathbf I}),
\label{eq:masked_hint_feature}
\end{equation}
where $\mathcal E_{\mathrm{geo}}^{1:b_{1}}$ encodes up to block $b_1$, and $\mathbf V^{b_1}$ is the validity mask aligned to its feature resolution.

Let $\mathbf F^1$ denote the clean Level~1 features of the ground-truth multi-view images. Given Gaussian noise $\boldsymbol{\epsilon}\sim\mathcal{N}(\mathbf{0},\mathbf{I})$ and $t\sim\mathcal{U}(0,1)$, our multi-view flow-matching model~\citep{flowmatching2023} adopts the linear probability path $\mathbf{X}_t=(1-t)\mathbf{F}^1+t\boldsymbol{\epsilon}$ with target velocity $\mathbf{u}_t=\boldsymbol{\epsilon}-\mathbf{F}^1$. The denoiser jointly predicts the velocity of all view tokens via:
\begin{equation}
    \hat{\mathbf{u}}_t
    =v_{\theta}\left(\mathbf{X}_t,t;\overline{\mathbf F},\boldsymbol\Gamma,\mathbf F^{\mathrm W},\mathbf D^{\mathrm{proj}},\mathbf V\right),
    \label{eq:l1_velocity}
\end{equation}
where $\boldsymbol\Gamma$ denotes camera Plücker-ray embeddings, $\mathbf F^{\mathrm W}$ represents Wan2.2 features, and $\mathbf D^{\mathrm{proj}}$ and $\mathbf V=[\mathbf 1_{\mathcal C},\mathbf V_{\mathcal T}]$ refer to projected depth guidance and its validity mask. Under these fixed conditions, we integrate the predicted velocity field from $t = 1$ to $t = 0$ to transform Gaussian noise into the Level-1 features $\hat{\mathbf{F}^1}$. These features then drive a conditional cascade, alongside context features and camera embeddings, to synthesize the Level-0 features $\hat{\mathbf{F}^0}$:
\begin{equation}
    \hat{\mathbf{F}^0}
    =\mathcal S_{\phi}(
    \boldsymbol\epsilon\mid\hat{\mathbf{F}^1},
    \mathbf F_{\mathcal C}^{0},\boldsymbol\Gamma),
    \label{eq:cascade_generation}
\end{equation}
where $\mathcal S_{\phi}$ denotes the cascade sampler initialized from Gaussian noise $\boldsymbol\epsilon$. Subsequently, the remaining frozen DA3 blocks forward-process $\hat{\mathbf{F}^1}$ to compute $(\hat{\mathbf{F}^2},\hat{\mathbf{F}^3})=\mathcal E_{\mathrm{geo}}^{b_1+1:b_3}(\hat{\mathbf{F}^1})$. Specialized RGB and geometry heads then utilize the multi-scale hierarchy $\hat{\mathcal{F}}=\{\hat{\mathbf{F}^i}\}_{i=0}^{3}$ to decode target appearance $\hat{\mathbf I_{\mathcal T}}=\mathcal D_{\mathrm{rgb}}(\hat{\mathcal F})_{\mathcal T}$ and target geometry $\hat{\mathbf G_{\mathcal T}}=\mathcal D_{\mathrm{geo}}(\hat{\mathcal F})_{\mathcal T}$.

\boldparagraph{Video-Prior Injection}
To complement GLD's geometric representation with learned visual priors, we use a frozen Wan2.2 VACE model~\citep{wan2025,vace2025}. Rather than sampling a complete video, we extract its intermediate features once and reuse them throughout geometric denoising. Further details of Wan2.2 VACE are provided in Appendix~\ref{app:wan_vace}.

Specifically, its feature extractor processes the RGB sequence $\overline{\mathbf I}$. Upon normalization and resizing, we inject Gaussian noise ($\sigma = 0.1$) exclusively into valid target projections, producing $\widetilde{\mathbf I}$ while leaving context RGB unperturbed. We encode $\widetilde{\mathbf I}$ with the frozen video VAE to obtain $\mathbf z_{\mathrm{proxy}}=E_{\mathrm{VAE}}(\widetilde{\mathbf I})$ and construct $\mathbf x_t=(1-\sigma_t)\mathbf z_{\mathrm{proxy}}+\sigma_t\boldsymbol\epsilon$, where $\boldsymbol\epsilon\sim\mathcal N(0,\mathbf I)$ and $\sigma_t$ is the noise level associated with the Wan timestep $t$. Using $\mathbf x_t$ as the backbone input and $[\mathbf z_{\mathrm{proxy}},\mathbf V]$ as the VACE condition, we extract features from blocks $(k_0,k_1,k_2,k_3)=(0,5,10,15)$ in a single forward pass:
\begin{equation}
\mathbf F^{\mathrm W,\ell}=\mathcal W^{0:k_\ell}\!\left(
\mathbf x_t,t;\mathcal E_{\mathrm{VACE}}([\mathbf z_{\mathrm{proxy}},\mathbf V])\right),
\label{eq:wan_features}
\end{equation}
where $\ell\in\{0,1,2,3\}$, $\mathcal E_{\mathrm{VACE}}$ denotes the VACE conditioning unit, $\mathcal W^{0:k_\ell}$ denotes the Wan backbone up to the selected block, and $\mathbf F^{\mathrm W}$ collectively denotes the extracted features.

To transfer these video features into the geometric latent space, a convolutional module $\mathcal A_\ell$ matches each feature $\mathbf F^{\mathrm W,\ell}$ to the spatial resolution and channel dimension of its corresponding injection layer. A ControlNet-style branch~\citep{controlnet2023} processes the aligned features under the same diffusion timestep and camera conditions as the main denoiser, then supplies an additive residual:
\begin{align}
\mathbf c_{\ell+1}&=\mathcal B_\ell^{\mathrm{ctrl}}(\mathbf c_\ell+\mathcal A_\ell(\mathbf F^{\mathrm W,\ell});t,\boldsymbol\Gamma),\label{eq:control_branch}\\
\mathbf h_{\ell+1}&=\mathcal B_\ell(\mathbf h_\ell;t,\boldsymbol\Gamma)+\mathcal Z_\ell(\mathbf c_{\ell+1}),\label{eq:control_injection}
\end{align}
where $\mathbf c_\ell$ and $\mathbf h_\ell$ are the control and main-branch features. The zero-initialized projections $\mathcal Z_\ell$ ensure that the auxiliary branch keeps the pretrained denoiser initially unmodified. During training, these residuals learn to adapt frozen video features for geometric generation, boosting appearance fidelity and completion while preserving explicit camera control.

\boldparagraph{Loss Function and Training}
We supervise both latent generation and decoded target predictions. The Level~1 denoiser uses a weighted flow-matching objective: 
\begin{equation}
\mathcal L_{\mathrm{FM}}^1=\mathbb E_{t,\boldsymbol\epsilon}\Big[
\sum_{i\in\mathcal C\cup\mathcal T}\alpha_i\Big\|
v_\theta\!\left(\mathbf X_t,t;
\overline{\mathbf F},\boldsymbol\Gamma,\mathbf F^{\mathrm W},\mathbf D^{\mathrm{proj}},\mathbf V\right)_i-\mathbf u_{t,i}\Big\|_2^2\Big],
\label{eq:l1_flow_loss}
\end{equation}
where $\alpha_i=0.25$ for context views and $1$ for target views, prioritizing novel-view synthesis while retaining context reconstruction. The Level~0 cascade uses an analogous objective $\mathcal L_{\mathrm{FM}}^0$.

For decoded target RGB, $\mathcal L_{\mathrm{rgb}}$ represents the mean valid-pixel $\ell_1$ reconstruction error, whereas $\mathcal L_{\mathrm{hf}}$ applies the same loss to high-gradient regions. Following DA3~\citep{da3_2025}, we estimate a camera-center similarity transform and leverage its scale $s^\star$ to align predicted depth, yielding $\mathcal L_{\mathrm{depth}}$ as the mean valid-pixel $\ell_1$ error between $s^\star\hat{\mathbf D_{\mathcal T}}$ and the reference depth. The full objective is
\begin{equation}
\mathcal L=\lambda_1\mathcal L_{\mathrm{FM}}^1+\lambda_0\mathcal L_{\mathrm{FM}}^0
+\lambda_{\mathrm{rgb}}\mathcal L_{\mathrm{rgb}}
+\lambda_{\mathrm{hf}}\mathcal L_{\mathrm{hf}}
+\lambda_{\mathrm{depth}}\mathcal L_{\mathrm{depth}}.
\label{eq:complete_training_loss}
\end{equation}

\subsection{Persistent Spatial Memory for Long-Sequence Consistency}
\label{sec:method_memory}
\boldparagraph{Spatial Memory Aggregation}
To maintain multi-round scene consistency, GeoVerse incrementally updates a colored point-cloud memory $\mathbf M^r$ in a shared coordinate system~\citep{viewcrafter2024,gen3c2025}. Context observations initialize the memory as $\mathbf M^0 = \mathbf P^0$ using DA3-predicted~\citep{da3_2025} or metric depth if available. For generated rounds, predicted depth is scale-aligned via $\widetilde{\mathbf D}_i^r = s_r^\star \hat{\mathbf D_i^r}$ using a camera-center similarity transform. Back-projecting valid pixels with $\widetilde{\mathbf D}_i^r$ and camera parameters $\mathbf C_i$ constructs a point set $\mathbf P^r$ containing 3D positions, colors, and confidence weights. Fusing each new point set $\mathbf P^r$ into $\mathbf M^{r-1}$ produces $\mathbf M^r$, placing accumulated content in a unified coordinate frame while mitigating scale drift.

\boldparagraph{Target-Aligned Projection}
For each target camera $\mathbf C_j$ ($j\in\mathcal T$), depth-aware point splatting renders the spatial memory $\mathbf M^r$ into RGB-D guidance $(\mathbf I_j^{\mathrm{proj}},\mathbf D_j^{\mathrm{proj}})$ and a validity mask $\mathbf V_j$, accounting for point confidence and z-buffer visibility. The mask $\mathbf V_j$ equals one for pixels supported by valid point projections and zero elsewhere. These projected hints and validity mask are subsequently encoded into target-side conditioning features via Eq.~(\ref{eq:masked_hint_feature}). Crucially, valid projections anchor known scene structures, while masked regions guide the generative prior to synthesize unobserved areas.

\begin{table}[t]
    \centering
    \caption{\textbf{Quantitative comparison across multiple datasets} in terms of visual quality, geometric accuracy, and average inference time. \textbf{Bold} and \underline{underline} denote the best and second-best results.}
    \label{tab:comparison}
    \scriptsize
    \setlength{\tabcolsep}{4.0pt}
    \resizebox{\textwidth}{!}{%
    \begin{tabular}{@{\hspace{5pt}}c@{\hspace{8pt}}lccc ccccc c}
        \toprule
         & \multirow{2}{*}{Method} & \multicolumn{3}{c}{2D Metrics} & \multicolumn{5}{c}{3D Metrics} & Efficiency \\
        \cmidrule(lr){3-5}\cmidrule(lr){6-10}\cmidrule(lr){11-11}
        & & PSNR$\uparrow$ & SSIM$\uparrow$ & LPIPS$\downarrow$ & ATE$\downarrow$ & RPE$_{\mathrm r}\downarrow$ & RPE$_{\mathrm t}\downarrow$ & Reproj.$\downarrow$ & MEt3R$\downarrow$ & Time (s)$\downarrow$ \\
        \midrule
        \multirow{8}{*}{\rotatebox[origin=c]{90}{\textbf{DL3DV}}} & ViewCrafter & 15.98 & 0.516 & 0.494 & 0.367 & 10.214 & 0.728 & 0.738 & 0.313 & 286.75 \\
         & NeoVerse & 12.14 & 0.353 & 0.622 & 0.164 & 7.934 & 0.382 & \textbf{0.638} & 0.278 & 249.19 \\
         & GEN3C & 17.06 & 0.561 & 0.435 & 0.114 & 5.788 & 0.238 & 0.692 & 0.301 & 343.10 \\
         & MVGenMaster & 17.28 & \underline{0.573} & 0.384 & 0.098 & 5.235 & 0.205 & 0.669 & 0.282 & 37.64 \\
         & Matrix3D & 13.47 & 0.416 & 0.494 & 0.153 & 6.127 & 0.323 & 0.715 & 0.288 & 53.39 \\
         & CAMEO & 11.14 & 0.383 & 0.677 & 0.906 & 44.157 & 1.987 & 0.887 & 0.425 & \underline{14.78} \\
         & GLD & \underline{17.38} & 0.546 & \underline{0.383} & \underline{0.058} & \underline{1.538} & \underline{0.125} & 0.652 & \underline{0.262} & 169.29 \\
         & \textbf{Ours} & \textbf{19.61} & \textbf{0.598} & \textbf{0.339} & \textbf{0.028} & \textbf{0.942} & \textbf{0.060} & \underline{0.649} & \textbf{0.261} & \textbf{9.24} \\
        \midrule
        \multirow{8}{*}{\rotatebox[origin=c]{90}{\textbf{RealEstate10K}}} & ViewCrafter & 16.84 & 0.646 & 0.407 & 0.091 & 1.715 & 0.160 & 0.685 & 0.208 & 287.87 \\
         & NeoVerse & 11.89 & 0.450 & 0.598 & 0.044 & 0.858 & 0.079 & 0.728 & 0.184 & 250.05 \\
         & GEN3C & 18.51 & 0.712 & 0.324 & 0.055 & \underline{0.512} & \textbf{0.060} & 0.686 & \underline{0.182} & 314.18 \\
         & MVGenMaster & \underline{20.54} & \underline{0.740} & \underline{0.295} & 0.039 & 0.991 & 0.067 & 0.653 & 0.196 & 26.78 \\
         & Matrix3D & 15.71 & 0.562 & 0.408 & 0.041 & 0.688 & 0.072 & 0.671 & 0.218 & 53.39 \\
         & CAMEO & 13.68 & 0.515 & 0.529 & 0.140 & 4.853 & 0.329 & 0.771 & 0.303 & \underline{14.28} \\
         & GLD & 19.66 & 0.709 & 0.299 & \underline{0.037} & 1.338 & 0.067 & \textbf{0.609} & 0.183 & 163.50 \\
         & \textbf{Ours} & \textbf{21.29} & \textbf{0.756} & \textbf{0.275} & \textbf{0.035} & \textbf{0.437} & \underline{0.065} & \underline{0.631} & \textbf{0.179} & \textbf{9.29} \\
        \midrule
        \multirow{8}{*}{\rotatebox[origin=c]{90}{\textbf{Mip-NeRF 360}}} & ViewCrafter & 15.84 & 0.417 & 0.553 & 0.446 & 19.046 & 1.102 & 0.706 & 0.346 & 295.81 \\
         & NeoVerse & 12.17 & 0.297 & 0.652 & 0.394 & 10.574 & 0.584 & 0.626 & 0.301 & 204.14 \\
         & GEN3C & 17.13 & \underline{0.465} & 0.476 & 0.420 & 11.549 & 0.679 & 0.674 & 0.304 & 339.81 \\
         & MVGenMaster & 16.75 & 0.446 & 0.427 & \underline{0.093} & 2.507 & 0.217 & 0.645 & 0.284 & 24.71 \\
         & Matrix3D & 15.06 & 0.360 & 0.489 & 0.118 & 2.819 & \underline{0.197} & 0.658 & 0.287 & 52.93 \\
         & CAMEO & 10.82 & 0.275 & 0.689 & 0.964 & 33.244 & 2.183 & 0.885 & 0.436 & \underline{14.95} \\
         & GLD & \underline{17.57} & 0.440 & \underline{0.406} & 0.105 & \underline{2.337} & 0.202 & \underline{0.621} & \underline{0.251} & 156.01 \\
         & \textbf{Ours} & \textbf{20.02} & \textbf{0.532} & \textbf{0.351} & \textbf{0.071} & \textbf{1.629} & \textbf{0.168} & \textbf{0.594} & \textbf{0.244} & \textbf{9.12} \\
        \midrule
        \multirow{5}{*}{\rotatebox[origin=c]{90}{\textbf{ScanNetV2}}} & ViewCrafter & 12.82 & 0.654 & 0.599 & 0.068 & 2.861 & 0.101 & 0.658 & 0.126 & 515.81 \\
         & NeoVerse & 10.53 & 0.422 & 0.671 & \underline{0.010} & 0.804 & 0.022 & \textbf{0.494} & \underline{0.100} & 605.04 \\
         & GEN3C & 12.54 & 0.548 & 0.567 & 0.058 & 2.478 & 0.071 & 0.665 & 0.169 & 1257.07 \\
         & GLD & \underline{15.33} & \underline{0.664} & \underline{0.491} & 0.011 & \underline{0.396} & \underline{0.019} & 0.598 & 0.108 & \underline{476.27} \\
         & \textbf{Ours} & \textbf{19.65} & \textbf{0.781} & \textbf{0.398} & \textbf{0.009} & \textbf{0.233} & \textbf{0.016} & \underline{0.589} & \textbf{0.088} & \textbf{33.11} \\
        \bottomrule
    \end{tabular}}
\end{table}

\subsection{Few-Step Inference via Reflow Distillation}
\label{sec:method_efficiency}
\boldparagraph{Hierarchical Few-Step Sampling}
We allocate three denoising steps to Level-1 for primary multi-view generation and one step to the frozen Level-0 cascade for shallow feature recovery. This budget allocation is empirically grounded in Table~\ref{tab:fewstep_analysis} and Fig.~\ref{fig:fewstep_qualitative}: three Level-1 steps recover noticeably finer details than a single step, whereas a single L0 cascade step suffices to match the quality of 49 steps. Consequently, distillation is applied solely to the Level-1 denoiser, leaving the cascade and decoding heads unchanged.

\boldparagraph{Level~1 Reflow Distillation} To achieve Level-1 generation within just three steps, we initialize a student model from the multi-step teacher and perform piecewise reflow~\citep{yan2024perflow} over three time intervals bounded by $1=\tau_0>\tau_1>\tau_2>\tau_3=0$. For notation brevity, the conditioning inputs from Eq.~(\ref{eq:l1_velocity}) remain fixed throughout each trajectory and are omitted below. For a given interval $a$, the starting state is constructed by corrupting clean ground-truth features $\mathbf F^1$ with Gaussian noise: $\mathbf X_a^+=(1-\tau_a)\mathbf F^1+\tau_a\boldsymbol\epsilon$. The frozen teacher then integrates the ODE from $\tau_a$ to $\tau_{a+1}$ to obtain the target endpoint $\mathbf X_a^-$. The effective velocity vector connecting these endpoints is computed as $\mathbf u_a^{\mathrm T}=(\mathbf X_a^+-\mathbf X_a^-)/(\tau_a-\tau_{a+1})$. By learning to match these straight-line endpoint displacements, the student effectively shortcuts multiple teacher evaluations into a single update per interval.

For any timestep $t\sim\mathcal U(\tau_{a+1},\tau_a)$, we form the linearly interpolated state $\widetilde{\mathbf X}_t=\eta\mathbf X_a^++(1-\eta)\mathbf X_a^-$, where $\eta=(t-\tau_{a+1})/(\tau_a-\tau_{a+1})$. The student learns this constant reflow velocity via
\begin{equation}
\mathcal L_{\mathrm{PRF}}^1=\mathbb E\Big[\sum_{j\in\mathcal C\cup\mathcal T}\alpha_j\big\|v_{\theta_{\mathrm S}}(\widetilde{\mathbf X}_t,t)_j-\mathbf u_{a,j}^{\mathrm T}\big\|_2^2\Big],
\label{eq:piecewise_reflow_loss}
\end{equation}
where the expectation is taken over training samples, noise vectors, intervals, and timesteps, and $\alpha_j$ retains the view-dependent weights from Sec.~\ref{sec:method_prior}. To stabilize optimization, we regularize the student with the original flow-matching objective: $\mathcal L_{\mathrm{fast}}^1=\rho\mathcal L_{\mathrm{PRF}}^1+(1-\rho)\mathcal L_{\mathrm{FM}}^1$. During inference, a single Euler step per interval traverses the trajectory before the cascade recovers Level-0 features.

\section{Experiments}
\label{sec:exp}

\begin{figure}[t]
    \centering
    \includegraphics[width=0.98\textwidth]{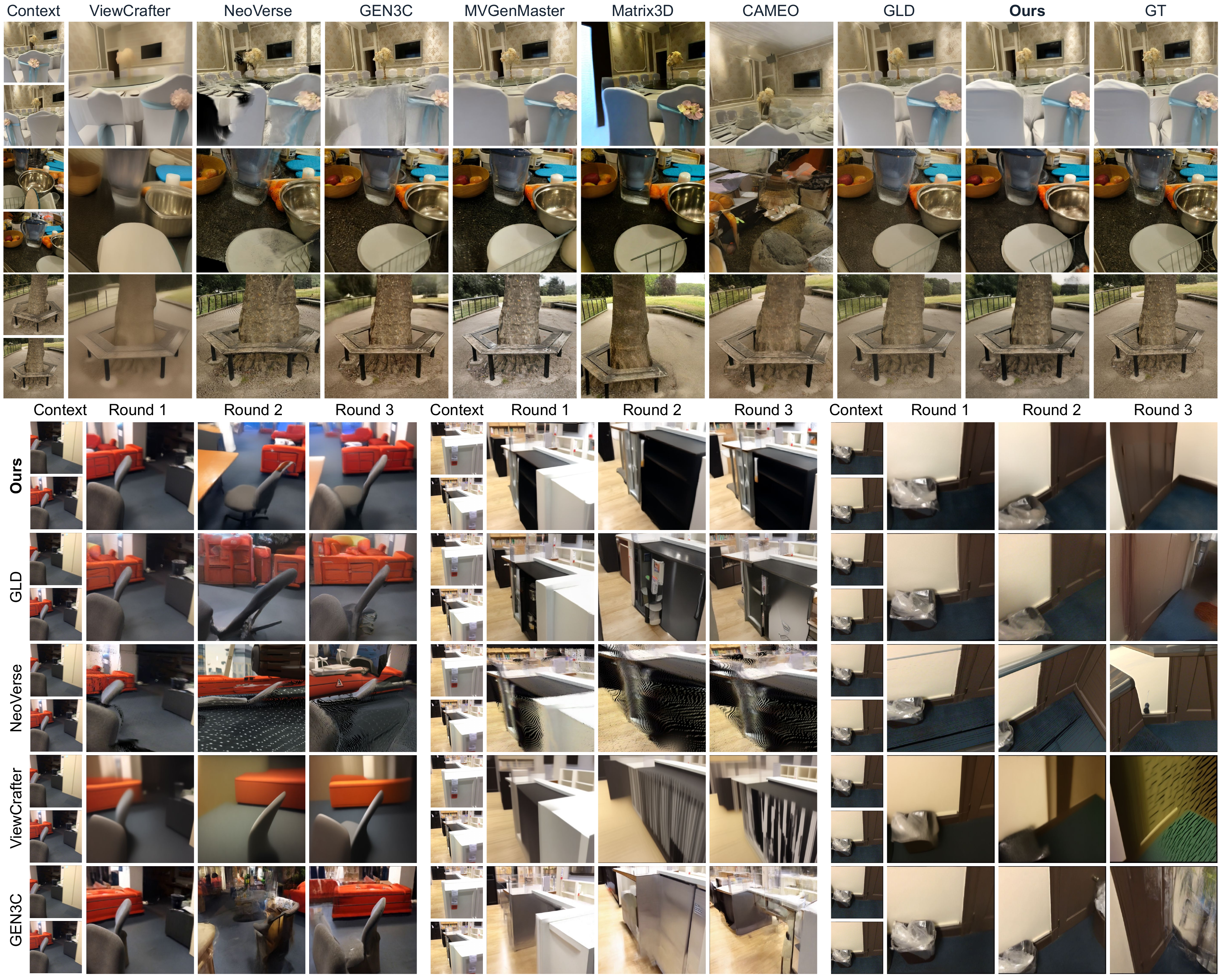}
     \vspace{-4pt}
     \caption{\textbf{Qualitative comparisons across multiple datasets.} Top: comparisons of results from a single inference pass. Bottom: comparisons of long-sequence generation over three rounds.}
    \label{fig:qualitative_comparison}
    \vspace{-8pt}
\end{figure}

\subsection{Experimental Setup}

\paragraph{Datasets and Metrics.}
We evaluate on two in-domain benchmarks, RealEstate10K~\citep{realestate10k2018} and DL3DV~\citep{dl3dv2024}, and two out-of-domain benchmarks, Mip-NeRF~360~\citep{mipnerf3602022} and ScanNetv2~\citep{scannet2017}, with ScanNetv2 used for long-sequence evaluation.
Specifically, each generation round uses $N_{\mathcal C}=2$ context views and $N_{\mathcal T}=6$ target views.
We report PSNR, SSIM~\citep{ssim2004}, and LPIPS~\citep{lpips2018} for visual quality, ATE and relative pose errors ($\mathrm{RPE}_{\mathrm r}$/$\mathrm{RPE}_{\mathrm t}$)~\citep{sturm2012rgbd} from VGGT-estimated poses~\citep{vggt2025} for target-camera fidelity, reprojection error and MEt3R~\citep{met3r2025} for cross-view geometric and feature consistency, and average inference time for efficiency.

\paragraph{Baselines.}
We compare with ViewCrafter~\citep{viewcrafter2024}, NeoVerse~\citep{neoverse2026}, GEN3C~\citep{gen3c2025}, MVGenMaster~\citep{mvgenmaster2025}, Matrix3D~\citep{matrix3d2025}, CAMEO~\citep{cameo2025}, and GLD~\citep{gld2026}, covering diverse approaches to generative novel-view synthesis.
For long-sequence evaluation, we compare with ViewCrafter, NeoVerse, GEN3C, and GLD across successive rounds to assess visual quality and cross-round consistency.

\paragraph{Implementation Details.}
Starting from a pretrained GLD~\citep{gld2026}, GeoVerse is trained on 15 real and synthetic multi-view datasets using DA3-Base~\citep{da3_2025} at $504\times504$ resolution, with eight ordered views per sample (one to four selected as context views). Training proceeds in two stages: a 300k-iteration adaptation phase with a learning rate of $3\times10^{-5}$, followed by up to 50k iterations of reflow distillation at $1\times10^{-5}$. Both stages are executed on 32 NVIDIA A800 GPUs using AdamW~\citep{adamw2019} with a global batch size of 32. Adaptation hyperparameters are set to $(\beta_1,\beta_2)=(0.9,0.95)$, gradient-norm clipping at 1.0, and an EMA decay of 0.9995. Further details are provided in Appendix~\ref{app:training_data}.

\subsection{Comparison}

\begin{table}[t]
    \centering
    \caption{\textbf{Quantitative comparisons on DL3DV~\citep{dl3dv2024} with different denoising budgets.} Three L1 updates and one Cascade L0 update provide the best overall trade-off between synthesis quality and inference speed among the compared configurations.}
    \vspace{-4pt}
    \label{tab:fewstep_analysis}
    \scriptsize
    \setlength{\tabcolsep}{3.0pt}
    \resizebox{\textwidth}{!}{%
    \begin{tabular}{r rrrrrr rrrrrr}
        \toprule
        \multirow{2}{*}{Steps}
        & \multicolumn{6}{c}{\textbf{L1 sweep (Cascade L0 fixed to 1)}}
        & \multicolumn{6}{c}{\textbf{Cascade L0 sweep (L1 fixed to 3)}} \\
        \cmidrule(lr){2-7}\cmidrule(lr){8-13}
        & Time (s)$\downarrow$ & PSNR$\uparrow$ & SSIM$\uparrow$ & LPIPS$\downarrow$ & ATE$\downarrow$ & MEt3R$\downarrow$
        & Time (s)$\downarrow$ & PSNR$\uparrow$ & SSIM$\uparrow$ & LPIPS$\downarrow$ & ATE$\downarrow$ & MEt3R$\downarrow$ \\
        \midrule
        49 & 80.04 & 17.72 & 0.535 & 0.376 & 0.037 & 0.290 & 52.55 & 19.48 & 0.597 & 0.342 & 0.025 & 0.260 \\
        24 & 41.33 & 17.87 & 0.541 & 0.370 & 0.035 & 0.286 & 29.88 & 19.50 & 0.598 & 0.342 & 0.027 & 0.261 \\
        12 & 23.02 & 18.15 & 0.550 & 0.362 & 0.033 & 0.282 & 19.29 & 19.53 & 0.599 & 0.342 & 0.029 & 0.261 \\
        6 & 13.84 & 18.61 & 0.567 & 0.349 & 0.032 & 0.272 & 13.75 & 19.60 & 0.600 & 0.342 & 0.031 & 0.271 \\
        3 & \textbf{9.24} & \textbf{19.61} & \textbf{0.598} & \textbf{0.339} & \textbf{0.028} & \textbf{0.261} & 11.02 & 19.66 & 0.601 & 0.341 & 0.029 & 0.261 \\
        1 & 6.07 & 20.55 & 0.630 & 0.367 & 0.028 & 0.269 & \textbf{9.24} & \textbf{19.61} & \textbf{0.598} & \textbf{0.339} & \textbf{0.028} & \textbf{0.261} \\
        \bottomrule
    \end{tabular}}
    \vspace{-12pt}
\end{table}

\begin{figure}[t]
    \centering
    
    \includegraphics[width=\textwidth]
    {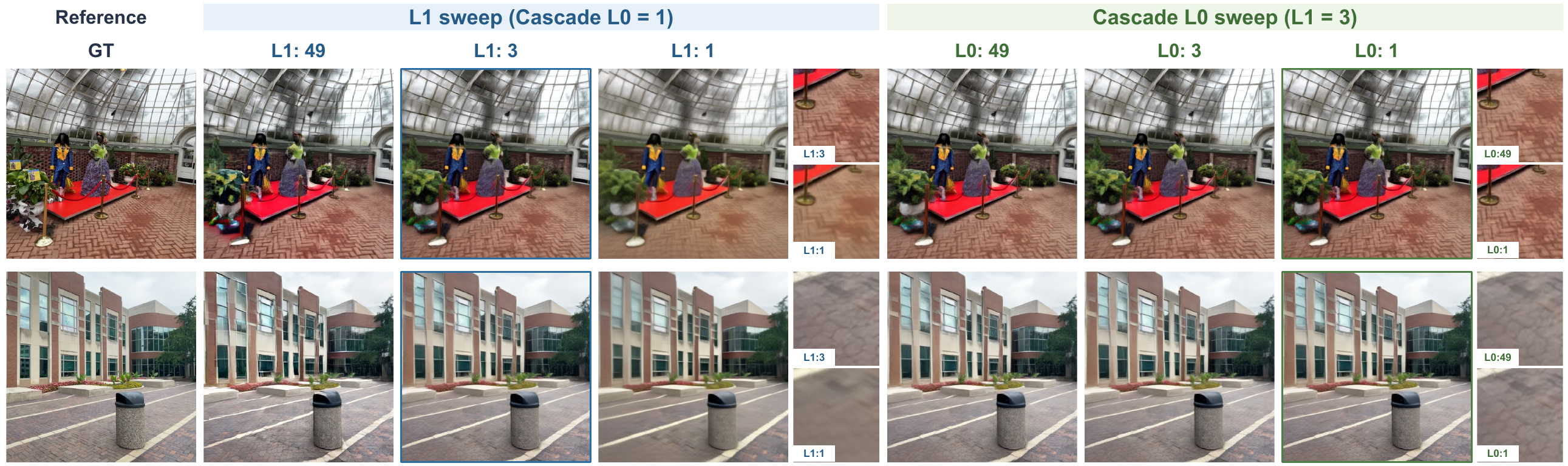}
    \caption{\textbf{Qualitative comparisons on DL3DV~\citep{dl3dv2024} with different denoising budgets.} One L1 update produces blurrier fine details than three L1 updates, while one Cascade L0 update preserves detail comparable to 49 Cascade L0 updates.}
    \vspace{-8pt}
    \label{fig:fewstep_qualitative}
    \vspace{-4pt}
\end{figure}

\paragraph{Visual Quality.}
GeoVerse achieves state-of-the-art performance across all four benchmarks in Tab.~\ref{tab:comparison}, surpassing GLD~\citep{gld2026} in PSNR by 2.23~dB on DL3DV and 2.45~dB on Mip-NeRF~360. These gains translate into clear visual improvements (Fig.~\ref{fig:qualitative_comparison}), effectively suppressing smearing and ghosting while recovering crisp object boundaries and faithful surface colors relative to context views. Unlike baselines that introduce misplaced structures or color shifts, GeoVerse preserves scene identity during synthesis. On long-sequence ScanNetv2 testing across three rounds, GeoVerse boosts PSNR from 15.33~dB (GLD) to 19.65~dB, maintaining temporal stability across large indoor structures.

\paragraph{Geometric Consistency.}
GeoVerse lowers ATE on Mip-NeRF~360 from 0.105 (GLD~\citep{gld2026}) to 0.071, representing a 32.4\% improvement. As reflected in Tab.~\ref{tab:comparison} and Fig.~\ref{fig:qualitative_comparison}, lower pose errors manifest as enhanced 3D structural fidelity, preserving intricate geometry like outdoor furniture without distortion. Over long sequences, key scene elements retain cross-round consistency, avoiding the structural repetitions and layout shifts prevalent in baselines. This validates the role of persistent spatial memory in anchoring new updates to accumulated context. While performance varies slightly across specific metrics, where some baselines achieve lower reprojection or translation errors on individual datasets, GeoVerse maintains significantly better global scene coherence.

\paragraph{Inference Efficiency.}
GeoVerse synthesizes six target views in roughly nine seconds across the three main benchmarks, operating about $18\times$ faster than GLD~\citep{gld2026}. This acceleration stems from using only four geometric denoising steps paired with single-pass video feature extraction. Even under this constrained sampling budget, generated outputs maintain crisp object contours and recognizable 3D geometry. This efficiency advantage naturally carries over to sequential view expansion, as each additional round employs the same compact pipeline. Per-pass runtimes for the main benchmarks and cumulative three-round times on ScanNetv2 are summarized in Tab.~\ref{tab:comparison}.

\subsection{Few-Step Analysis}
\label{sec:fewstep_analysis}

We analyze the denoising budget within GLD's hierarchical sampling framework~\citep{gld2026}, with Level-1 acceleration based on reflow~\citep{yan2024perflow}. Table~\ref{tab:fewstep_analysis} evaluates denoising budget trade offs on DL3DV~\citep{dl3dv2024}. Reducing Level-1 generation from three steps to one lowers runtime from 9.24 to 6.07~s, yet worsens LPIPS~\citep{lpips2018} from 0.339 to 0.367. As shown in Fig.~\ref{fig:fewstep_qualitative}, this setting causes a loss of fine brick textures and object boundaries despite higher PSNR, showing that PSNR alone fails to reflect perceptual quality. On the other hand, cutting Level~0 cascade steps from 49 to one preserves comparable local details with an LPIPS of 0.342 versus 0.339 at much lower latency. These complementary trends validate our $3+1$ budget allocation.

\subsection{Ablation Studies}
\label{sec:ablation_studies}

Table~\ref{tab:ablation_corrected} summarizes diagnostic ablations on ScanNetv2~\cite{scannet2017} across model components and supervision losses. (1)~Removing the Wan2.2~\cite{wan2025} video prior causes the most severe visual degradation, proving that video appearance priors are essential for synthesizing faithful content. (2)~Disabling spatial memory substantially degrades both visual fidelity and cross view alignment, highlighting the importance of accumulated scene context for multi round consistency. (3)~The Mixture of Transformers variant explores key value attention as an alternative to additive residual injection, though isolating its relative advantages requires comparison under matched training settings. (4)~Omitting RGB supervision leads to a moderate decline in visual quality, showing that direct appearance supervision complements the video prior. (5)~Removing depth supervision weakens overall geometric consistency and confirms the value of explicit structural constraints, though non uniform metric variations indicate nuanced geometric trade offs. 

\begin{table*}[t]
    \centering
    \caption{\textbf{Ablation study on ScanNetV2~\citep{scannet2017}.} We evaluate the contributions of key components, training losses, and alternative architectures for video-prior injection.}
    \vspace{-4pt}
    \label{tab:ablation_corrected}
    \scriptsize
    \setlength{\tabcolsep}{5pt}
    \resizebox{\textwidth}{!}{%
    \begin{tabular}{lccc ccccc c}
        \toprule
        \multirow{2}{*}{Variant} & \multicolumn{3}{c}{2D Metrics} & \multicolumn{5}{c}{3D Metrics} & Scaled time \\
        \cmidrule(lr){2-4}\cmidrule(lr){5-9}\cmidrule(lr){10-10}
        & PSNR$\uparrow$ & SSIM$\uparrow$ & LPIPS$\downarrow$
        & ATE$\downarrow$ & RPE$_{\mathrm r}\downarrow$ & RPE$_{\mathrm t}\downarrow$ & Reproj.$\downarrow$ & MEt3R$\downarrow$ & Reference (s)$\downarrow$ \\
        \midrule
        \textbf{GeoVerse} & 19.65 & 0.781 & 0.398 & 0.009 & 0.223 & 0.016 & 0.589 & 0.088 & 33.11 \\
        w/o Wan2.2 prior & 16.83 & 0.696 & 0.463 & 0.022 & 0.388 & 0.034 & 0.591 & 0.092 & 29.19 \\
        w/o spatial memory & 17.87 & 0.727 & 0.434 & 0.014 & 0.268 & 0.023 & 0.591 & 0.087 & 33.48 \\
        \midrule
        MoT video-KV & 19.24 & 0.769 & 0.403 & 0.009 & 0.238 & 0.016 & 0.587 & 0.089 & 36.29 \\
        w/o RGB loss & 18.58 & 0.756 & 0.415 & 0.014 & 0.360 & 0.023 & 0.581 & 0.091 & 33.21 \\
        w/o depth loss & 19.05 & 0.764 & 0.413 & 0.010 & 0.282 & 0.015 & 0.588 & 0.092 & 33.29 \\
        \bottomrule
    \end{tabular}}
    \vspace{-8
    pt}
\end{table*}

\section{Limitations}
\label{sec:limitations}

While GeoVerse achieves strong multi-view consistency and high inference efficiency, two main limitations remain. First, its peak visual fidelity is constrained by the appearance capacity of the DA3 feature space. By prioritizing speed through compact feature conditioning, GeoVerse may not fully match the fine texture richness of computationally heavy, full-scale video diffusion models. Second, multi-round consistency depends on both the underlying geometry backbone and synthesized RGB coherence. In textureless or complex regions, initial depth errors and generated appearance drift can accumulate through spatial memory updates, occasionally compromising cross-view alignment across long trajectories. Improving joint 3D geometric and photometric stability remains a key objective for future research.

\section{Conclusion}
\label{sec:conclusion}
We introduced GeoVerse, a framework designed to resolve the fundamental trade off between geometric consistency and generative completion in novel view synthesis. By embedding pretrained video appearance priors into a 3D geometric latent diffusion model, GeoVerse generates high fidelity views while maintaining rigid spatial structure. Our global spatial memory further prevents cumulative drift across extended trajectories by grounding ongoing synthesis in accumulated 3D scene context. Accelerated by a few step reflow distillation scheme, GeoVerse achieves over 17 times speedup compared to GLD while setting new state of the art benchmarks in visual quality and pose accuracy. By demonstrating the effectiveness of combining video learned priors with geometric latent spaces, GeoVerse establishes a robust foundation and offers valuable insights toward building multi view consistent video world models.
\section*{AI use statement}
\label{sec:statement}

In this work, we used generative AI tools for assisting with translation. We have not used generative AI tools for designing research methods and experiments, implementing methodologies, interpreting results, proposing or refining hypotheses, cleaning and reformatting datasets, or supporting qualitative and thematic data analysis, and generating synthetic datasets, proposing mathematical claims, providing key elements for proving mathematical claims, and assisting in writing proofs are not applicable to this work. Additionally, we used generative AI tools for summarizing or analyzing existing literature, and editing the manuscript to enhance readability. We have reviewed all AI-assisted work: translated or polished text was manually cross-checked sentence-by-sentence to ensure that the original intent remained uncompromised. We take responsibility for the final content of this work, including text, claims or artifacts produced with the aid of generative AI.

\bibliography{main}
\bibliographystyle{main}

\clearpage
\appendix
This supplementary material provides additional technical details and experimental results to complement the main paper.
Section~\ref{app:wan_vace} introduces Wan2.2 VACE, including its inference pipeline, pretraining scale, and use as a frozen feature extractor in GeoVerse.
Section~\ref{app:loss_details} defines the training losses and details their computation, including high-frequency weighting and depth-scale alignment.
Section~\ref{app:mot} describes the key/value-based MoT variant used in the architectural comparison.
Section~\ref{app:training_data} summarizes the training data, with dataset types, scene counts, and image counts.
Section~\ref{app:qualitative} presents additional qualitative comparisons for novel-view synthesis and successive view expansion.

\section{Wan2.2 VACE Details}
\label{app:wan_vace}
\paragraph{Video generation and conditioning.}
Wan is a video diffusion framework with a causal video VAE, a text encoder, and a diffusion Transformer~\citep{wan2025}. In the standard generation pipeline, the Transformer iteratively denoises a video latent under text and optional visual conditions, and the VAE decoder converts the final latent into RGB frames. Visual inputs requiring latent encoding are processed by the VAE encoder before entering the corresponding conditioning path. Wan2.2-A14B uses separate high- and low-noise experts for different portions of the denoising trajectory, with approximately 14 billion active parameters per step.

VACE~\citep{vace2025} adds a unified interface for reference images, source videos, and spatiotemporal masks. Its Video Condition Unit organizes these inputs for reference-guided generation, video editing, inpainting, outpainting, and temporal extension. The standard VACE pipeline tokenizes visual conditions through context encoding and introduces them into the video backbone through context adapters. Its training curriculum progresses from foundational completion tasks to multiple references and task combinations, followed by quality refinement. These components provide visual completion cues that complement the geometric representation used by GeoVerse.

\paragraph{Pretraining data scale.}
The Wan technical report describes pretraining on billions of images and videos~\citep{wan2025}. The Wan2.2 release reports 65.6\% more images and 83.2\% more videos than Wan2.1; these are relative increases, rather than disclosed absolute counts for the Wan2.2 corpus. VACE constructs task-specific conditions from filtered videos, including segmentation masks, reference crops, depth, pose, and optical flow~\citep{vace2025}. The available model card does not specify an absolute training-set size for the particular Wan2.2-VACE-Fun checkpoint used here. These external pretraining corpora are separate from GeoVerse's multi-view training data in Sec.~\ref{app:training_data}.

\paragraph{Feature extraction in GeoVerse.}
Our checkpoint is derived from the low-noise component of Wan2.2-VACE-Fun-A14B~\citep{wan2025}. We retain backbone blocks 0--15 and the corresponding VACE blocks, extracting 5,120-dimensional features at blocks 0, 5, 10, and 15. The frozen extractor runs once per multi-view prediction, and the geometric denoiser reuses these features across its denoising steps.

The two input branches share an RGB proxy assembled from observed context images and projected target images. After normalization and resizing, Gaussian perturbations with standard deviation 0.1 are added only to valid target projections; context RGB remains unchanged and invalid target regions are zero-filled. Target ground-truth RGB is not used to construct this proxy.

For the Wan backbone, the frozen video VAE encodes the proxy into a 16-channel latent $\mathbf z_{\mathrm{proxy}}=E_{\mathrm{VAE}}(\widetilde{\mathbf I})$, using the checkpoint's latent normalization. We then form $\mathbf x_t=(1-\sigma_t)\mathbf z_{\mathrm{proxy}}+\sigma_t\boldsymbol\epsilon$, with $\boldsymbol\epsilon\sim\mathcal N(0,\mathbf I)$, and apply the backbone's patch embedding. The extraction timestep is fixed at $t=500$, and its noise level $\sigma_t$ is obtained from the corresponding scheduler mapping. This latent-space perturbation is distinct from the 0.1 RGB perturbation above and is applied across the latent, rather than only in invalid regions.

For the VACE branch, we convert visibility $\mathbf V$ into the generation mask $\mathbf m=1-\mathbf V$. Following VACE's condition encoding~\citep{vace2025}, the inactive and reactive RGB inputs, $\widetilde{\mathbf I}\odot(1-\mathbf m)$ and $\widetilde{\mathbf I}\odot\mathbf m$, are separately VAE-encoded into two 16-channel latents. Each spatial $8\times8$ mask block is rearranged into 64 channels and temporally aligned to the latent grid. Concatenating the two latents and the packed mask produces the 96-channel VACE condition; the mask itself is not VAE-encoded. The VACE blocks combine the encoded condition with backbone tokens and supply control residuals to the corresponding Wan blocks. Unlike the backbone input, the condition latents receive no additional timestep-dependent diffusion noise. This single-pass feature extraction does not run a complete video-sampling trajectory or invoke the VAE decoder; its features are reused by the ControlNet-style adapters~\citep{controlnet2023} in the geometric denoiser.

\section{Loss Definitions and Computation}
\label{app:loss_details}
We expand the five loss terms in Eq.~(\ref{eq:complete_training_loss}). The formulas below describe a single multi-view training sample; losses are averaged over the mini-batch. Latent flow matching supervises context and target views, whereas decoded RGB and depth losses supervise target views only.

\paragraph{Two-level flow matching.}
For feature level $i\in\{0,1\}$, we sample $t\sim\mathcal U(0,1)$ and Gaussian noise with the same shape as the clean reference features $\mathbf F^i$. The noisy state is $\mathbf X_t^i=(1-t)\mathbf F^i+t\boldsymbol\epsilon$, and its velocity target is $\boldsymbol\epsilon-\mathbf F^i$. Writing $\hat{\mathbf u}_{t,j}^{\,i}$ for the predicted velocity of view $j$, we obtain
\begin{equation}
\mathcal L_{\mathrm{FM}}^i
=\mathbb E_{t,\boldsymbol\epsilon}\!\left[
\sum_{j\in\mathcal C\cup\mathcal T}\alpha_j
\left\|\hat{\mathbf u}_{t,j}^{\,i}-(\boldsymbol\epsilon_j-\mathbf F_j^i)\right\|_2^2
\right],\qquad i\in\{0,1\}.
\label{eq:supp_flow_loss}
\end{equation}
The squared norm sums errors over the feature grid and channels of each view. We use $\alpha_j=0.25$ for context views and $1$ for target views. At Level~1, the velocity is predicted by $v_\theta$ under the conditions in Eq.~(\ref{eq:l1_velocity}); at Level~0, it is predicted by the denoiser underlying the conditional cascade in Eq.~(\ref{eq:cascade_generation}).

\paragraph{RGB supervision.}
Let $\Omega_j^{\mathrm{rgb}}$ denote the pixels with valid reference RGB in target view $j$, and define the channel-averaged error $e_j(p)=\|\hat{\mathbf I_j}(p)-\mathbf I_j(p)\|_1/3$. The reconstruction loss is
\begin{equation}
\mathcal L_{\mathrm{rgb}}
=\frac{1}{N_{\mathrm{rgb}}}
\sum_{j\in\mathcal T}\sum_{p\in\Omega_j^{\mathrm{rgb}}}e_j(p).
\label{eq:supp_rgb_loss}
\end{equation}
Here, $N_{\mathrm{rgb}}=\max(1,\sum_{j\in\mathcal T}|\Omega_j^{\mathrm{rgb}}|)$ is the valid-pixel count clamped to at least one. Predictions and references use the same RGB normalization. These supervision pixels are distinct from the projection-visibility mask $\mathbf V$: valid reference pixels remain supervised even when they are not visible in the memory projection.

To further emphasize image boundaries and texture, we weight the same RGB error using reference-image gradients. We first convert reference RGB to grayscale and compute its gradient magnitude $g_j(p)$ using horizontal and vertical $3\times3$ Sobel filters. Let $q_{0.50}$ and $q_{0.98}$ be the corresponding gradient-magnitude quantiles over valid target pixels in the sample. We form $w_j(p)=\operatorname{clip}_{[0,1]}((g_j(p)-q_{0.50})/(q_{0.98}-q_{0.50}+\delta))$, with $\delta=10^{-6}$, and compute
\begin{equation}
\mathcal L_{\mathrm{hf}}
=\frac{1}{Z_{\mathrm{hf}}}
\sum_{j\in\mathcal T}\sum_{p\in\Omega_j^{\mathrm{rgb}}}w_j(p)e_j(p).
\label{eq:supp_hf_loss}
\end{equation}
The normalizer is $Z_{\mathrm{hf}}=\max(\delta,\sum_{j\in\mathcal T}\sum_{p\in\Omega_j^{\mathrm{rgb}}}w_j(p))$. The weights are computed from the reference images, so the model cannot reduce this term by suppressing its own image gradients. The loss is zero when all weights are zero.

\paragraph{Scale-aligned depth supervision.}
Following the camera-center alignment described in the main paper, let $\hat{\mathbf o_j}$ and $\mathbf o_j$ denote predicted and reference camera centers for views with valid camera estimates, indexed by $\mathcal J\subseteq\mathcal C\cup\mathcal T$. We fit a single similarity transform per sample:
\begin{equation}
(s^\star,\mathbf Q^\star,\mathbf b^\star)
=\underset{s>0,\,\mathbf Q\in\mathrm{SO}(3),\,\mathbf b\in\mathbb R^3}{\arg\min}
\sum_{j\in\mathcal J}
\left\|s\mathbf Q\hat{\mathbf o_j}+\mathbf b-\mathbf o_j\right\|_2^2.
\label{eq:supp_camera_alignment}
\end{equation}
We solve this least-squares problem by centering the camera centers and applying singular value decomposition to their cross-covariance. Only the recovered scale $s^\star$ is applied to camera-space depth; rotation and translation align the camera coordinate systems and do not enter the depth residual. The same scale is shared by all target views, rather than fitted separately to each depth map. With $\Omega_j^{\mathrm{depth}}$ denoting pixels whose reference depth is finite and positive, the loss is
\begin{equation}
\mathcal L_{\mathrm{depth}}
=\frac{1}{N_{\mathrm{depth}}}
\sum_{j\in\mathcal T}\sum_{p\in\Omega_j^{\mathrm{depth}}}
\bigl|s^\star\hat{\mathbf D_j}(p)-\mathbf D_j(p)\bigr|.
\label{eq:supp_depth_loss}
\end{equation}
Here, $N_{\mathrm{depth}}=\max(1,\sum_{j\in\mathcal T}|\Omega_j^{\mathrm{depth}}|)$ normalizes by the valid depth-pixel count.

\begin{table}[t!]
\centering
\caption{\textbf{Training datasets used by GeoVerse.} We summarize the data types, scene counts, and image counts of the real and synthetic multi-view datasets in our training mixture.}
\label{tab:supp_training_data}
\small
\setlength{\tabcolsep}{7pt}
\begin{tabular}{@{}llrr@{}}
\toprule
Dataset & Data type & \#Scenes & \#Images \\
\midrule
Aria Digital Twin~\citep{adt2023} & Real / digital twin & 188 & 232,115 \\
ARKitScenes~\citep{arkitscenes2021} & Real / RGB-D & 644 & 125,134 \\
DL3DV~\citep{dl3dv2024} & Real / multi-view video & 10,475 & 3,594,809 \\
MapFree~\citep{mapfree2022} & Real / multi-view video & 460 & 515,113 \\
ScanNet++ v2~\citep{scannetpp2023} & Real / RGB + scans & 954 & 1,032,198 \\
Waymo~\citep{waymo2020} & Real / driving & 3,990 & 790,405 \\
RealEstate10K~\citep{realestate10k2018} & Real / real-estate video & 66,033 & 8,832,823 \\
\midrule
GTA-SfM~\citep{gtasfm2019} & Synthetic / outdoor & 200 & 17,649 \\
Hypersim~\citep{hypersim2021} & Synthetic / indoor & 107 & 17,348 \\
MVSSynth~\citep{deepmvs2018} & Synthetic / driving & 120 & 12,000 \\
OmniWorld~\citep{omniworld2025} & Synthetic / diverse scenes & 4,486 & 794,596 \\
TartanAir~\citep{tartanair2020} & Synthetic / diverse scenes & 18 & 613,274 \\
TartanAir v2~\citep{tartanair2020} & Synthetic / diverse scenes & 74 & 7,135,955 \\
Virtual KITTI~\citep{vkitti2020} & Synthetic / driving & 100 & 42,520 \\
Spring~\citep{spring2023} & Synthetic / dynamic scenes & 37 & 5,000 \\
\midrule
Total & 15 datasets & 87,886 & 23,760,939 \\
\bottomrule
\end{tabular}
\end{table}

In each training iteration, we compute the latent velocity residuals, decode target RGB and depth from the predicted feature hierarchy, construct the RGB gradient weights and camera-based scale, and combine the resulting terms using the coefficients in Eq.~(\ref{eq:complete_training_loss}). The reflow objective in Sec.~\ref{sec:method_efficiency} is a separate distillation objective.

\section{MoT Variant for Video-Prior Injection}
\label{app:mot}
\paragraph{Shared-attention conditioning.}
Following the shared-attention design of Fast-WAM~\citep{yuan2026fastwam}, we couple video-prior tokens and geometric latent tokens through attention while retaining separate processing streams. At each coupled layer, modality-specific projections map the two representations into a common attention space. The geometric stream produces queries, keys, and values from its current denoising state, while the video stream supplies visual keys and values. We concatenate the video and geometric keys and values along the token dimension, allowing each geometric query to attend jointly to scene features and video priors. The resulting attention output passes through the geometric stream's output projection, residual connection, and feed-forward layers to update its tokens.

\paragraph{Feature reuse during denoising.}
The interaction is asymmetric: geometric tokens attend to the video stream, while video features remain independent of the evolving geometric state. We therefore compute and cache the video keys and values once, reusing them across denoising updates. Geometric queries, keys, and values are recomputed at each step. This variant introduces video information within the attention computation, whereas the ControlNet-style design injects it through an auxiliary residual branch.

\suppressfloats[t]
\section{Training Data and Statistics}
\label{app:training_data}
\paragraph{Dataset composition.}
Our training mixture contains 15 datasets spanning real indoor captures, outdoor videos, driving sequences, and synthetic environments. TartanAir and TartanAir v2 are listed separately as distinct sampling sources. Table~\ref{tab:supp_training_data} follows the dataset/type/count presentation of DA3~\citep{da3_2025}, but reports our own training-manifest and RGB-inventory statistics rather than DA3's differently selected subsets. MegaDepth is excluded from this mixture.

\paragraph{Sampling and geometric inputs.}
We sample dataset sources with probabilities proportional to the square root of their RGB frame counts, balancing coverage against the dominance of the largest collections. RealEstate10K and Waymo use Pi3-estimated poses and depth~\citep{pi3_2025}. For the remaining sources, available dataset geometry and Pi3-estimated geometry are selected at a nominal ratio of $2:1$; samples with missing or incomplete depth fall back to Pi3. Samples that fail loading or validity checks are resampled. The inventory counts therefore describe the available training pool, not the number of distinct images consumed by a completed run.

\section{Additional Qualitative Results}
\label{app:qualitative}
\subsection{Novel-View Synthesis}
\begin{figure}[!ht]
    \centering
    \includegraphics[width=\textwidth]{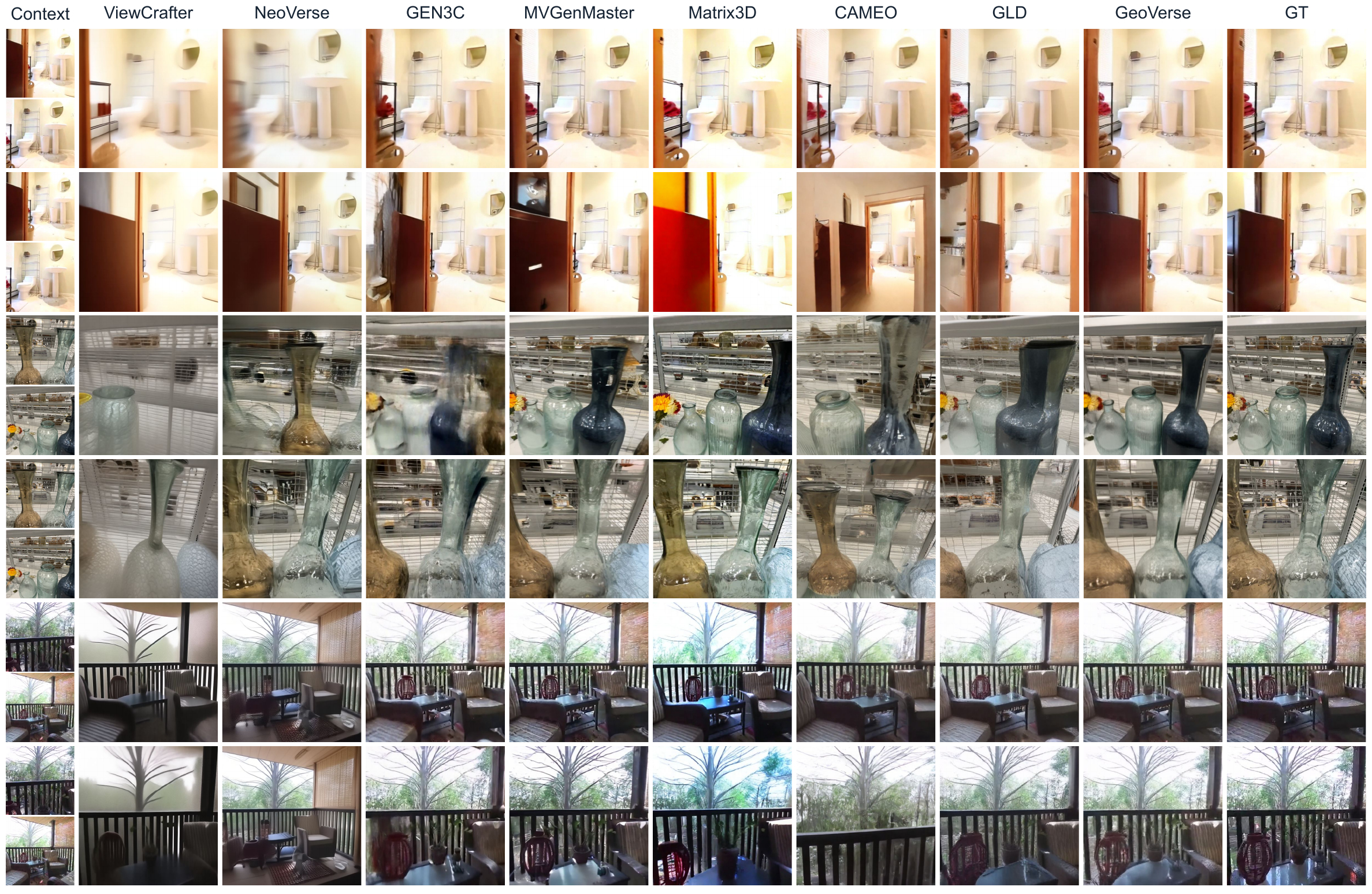}
    \caption{\textbf{Additional qualitative comparisons on RE10K and DL3DV.} Two target views are shown for each of three scenes, alongside context images and ground truth.}
    \label{fig:supp_qualitative}
\end{figure}

Figure~\ref{fig:supp_qualitative} extends the main-paper comparison with two target views from each of three scenes from RE10K datasets and DL3DV datasets.

\clearpage
\subsection{Successive View Expansion}
\label{app:long_sequence}
\begin{figure}[!ht]
    \centering
    \includegraphics[width=\textwidth]{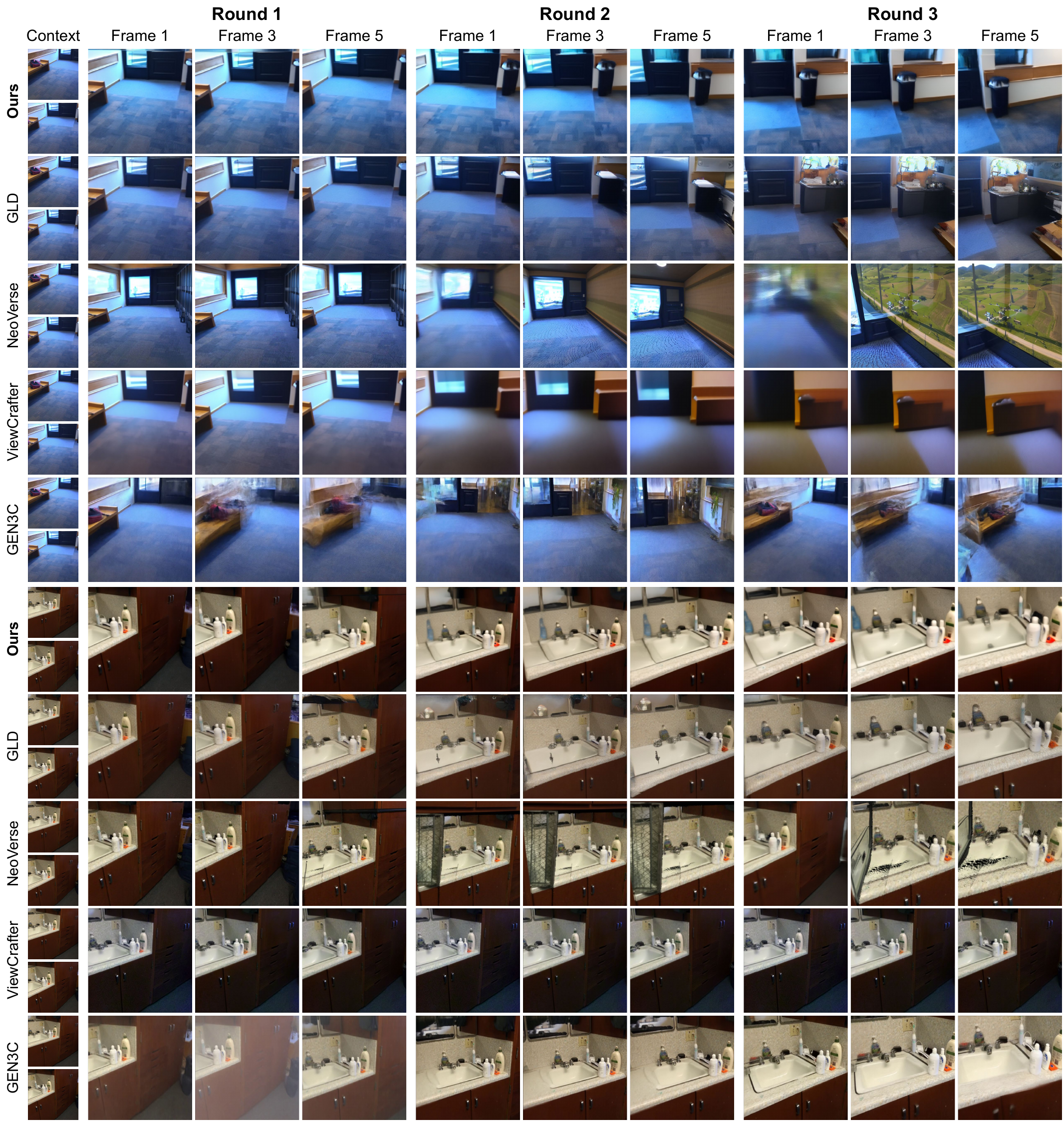}
    \caption{\textbf{Additional long-sequence comparisons on ScanNetv2.} Each row contains two context images followed by target frames 1, 3, and 5 from each of three rounds.}
    \label{fig:supp_long_sequence}
\end{figure}

Figure~\ref{fig:supp_long_sequence} presents three successive generation rounds on two ScanNetv2 scenes. We display target frames 1, 3, and 5 from each round to illustrate how scene appearance evolves as viewpoints expand.

\end{document}